\documentclass{article}
\usepackage{iclr2022_workshop,times}

\usepackage{microtype}
\usepackage{graphicx}
\usepackage{subfigure}
\usepackage{booktabs} 
\usepackage{amsmath}
\usepackage{amssymb}
\usepackage{mathtools}
\usepackage{amsthm}

\usepackage{amsmath,amsfonts,bm,amssymb,amsfonts,stmaryrd}

\renewcommand\vec{\bm}
\newcommand{\mat}[1]{{\bm{#1}}}
\def\Df{\mat{\mathcal{D}}}

\newcommand{\Fc}{\mathcal{F}}

\DeclareMathOperator{\face}{\trianglelefteqslant}

\newcommand{\constshf}[1]{\underline{#1}}

\def\eqref#1{equation~\ref{#1}}

\def\1{\bm{1}}

\DeclareMathAlphabet{\mathsfit}{\encodingdefault}{\sfdefault}{m}{sl}
\SetMathAlphabet{\mathsfit}{bold}{\encodingdefault}{\sfdefault}{bx}{n}

\theoremstyle{plain}
\newtheorem{theorem}{Theorem}[section]
\newtheorem{proposition}[theorem]{Proposition}

\theoremstyle{definition}
\newtheorem{definition}[theorem]{Definition}

\theoremstyle{remark}

\usepackage{hyperref}

\title{Graph Convolutional Networks from the Perspective of Sheaves and the Neural Tangent Kernel}

\author{Thomas Gebhart \\
Department of Computer Science\\
University of Minnesota \\
Minneapolis, MN 55455, USA \\
\texttt{gebhart@umn.edu} \\
}

\iclrfinalcopy 
\begin{document}

\maketitle

\begin{abstract}
Graph convolutional networks are a popular class of deep neural network algorithms which have shown success in a number of relational learning tasks. 
Despite their success, graph convolutional networks exhibit a number of peculiar features, including a bias towards learning oversmoothed and homophilic functions, which are not easily diagnosed due to the complex nature of these algorithms. 
We propose to bridge this gap in understanding by studying the neural tangent kernel of sheaf convolutional networks--a topological generalization of graph convolutional networks. 
To this end, we derive a parameterization of the neural tangent kernel for sheaf convolutional networks which separates the function into two parts: one driven by a forward diffusion process determined by the graph, and the other determined by the composite effect of nodes' activations on the output layer. 
This geometrically-focused derivation produces a number of immediate insights which we discuss in detail.
\end{abstract}

\section{Introduction}

Graph neural networks (GNNs) are a class of deep learning architectures which aim to learn functions over relationally-structured data. 
GNNs come in a variety of forms~\cite{xu2018powerful, kipf2016semi, velivckovic2022message} which generally compute, within each layer, nonlinear functions of message-passing operations that update signals at each node in the input graph according to some notion of locality.
The addition of deeper layers to these networks further propagates  messages outwards from each node's local neighborhood, turning local message-passing operations into global transformations. 
The general and flexible computational structure provided by GNNs has led to their achievement of state-of-the-art performance on tasks spanning a variety of application domains from social science to drug design~\cite{chen2018fastgcn, liao2018lanczosnet}. 

Despite these successes, GNNs face a number of practical and theoretical shortcomings. 
For example, deep GNNs are known to over-smooth input data leading to the learning of rather generic function on the input graph signals which results in poor performance~\cite{chen2020simple}. 
In addition, many types of GNNs are known to make strong assumptions about input graphs being homophilic--that connected nodes will be more similar to each other and will share more properties relative to other nodes in the network~\cite{zhu2020beyond}.

While architectural fixes for these types of problems have been proposed~\cite{zhu2020beyond, bodnar2022neural}, the layer-wise nature of GNNs, in combination with the representational flexibility of graph-structured data, leads to substantial complexity and makes an intuitive understanding of the function of these algorithms elusive.
Given these limitations, significant prior work has been devoted to these analytical tasks, resulting in a variety of approaches for characterizing the learning dynamics, bounding generalization performance, and formalizing the representational capacity of GNNs~\cite{du2019graph, xu2018powerful, xu2020neural, bodnar2022neural}. 

One approach for approximating both a formal and intuitive grasp on the behavior of neural networks in general is by analyzing their asymptotic behavior under gradient descent through the lens of the neural tangent kernel of the network~\cite{jacot2018neural, arora2019exact}. 
In particular, \citet{du2019graph} showed that the neural tangent kernel of graph convolutional networks (GCNs), a large and popular class of GNN architectures, can be described by a recursive relationship among the feature covariance of connected nodes. 
The authors were then able to use this interpretation to provide bounds on the asymptotic behavior of infinitely-wide GCNs and the function classes learnable by their studied architectures. 

In this paper, we offer an extension to the study of GCNs through the their tangent kernels by developing a geometrically-oriented reformulation of their neural tangent kernel.
We start by deriving the tangent kernel of a more general graph-convolutional architecture known as a sheaf convolutional network~\cite{hansen2020sheaf}. 
From the vantage point of this more general architecture, we gain the necessary perspective for reasoning about the shortcomings of GCNs through a kernel operator associated to sheaf convolutional networks. 
As we will observe, the resulting geometric reformulation is intuitive, providing insight into not only the functional behavior of GCNs, but also their idiosyncratic deficiencies.
This reformulation emphasizes the susceptibility of deep GCNs to oversmoothing and provides a framework through which to make spectral arguments about their performance under particular distributions of structure over the input graphs. 
We also observe a relationship between the graph neural tangent kernel, and GCNs by extension, to more traditional diffusion-based graph kernels, charting a course for future work in designing relational deep learning architectures whose tangent kernels approximate more exotic graph kernels. 

\subsection{Notation}
We denote vectors and matrices in bold script, respectively $\vec{x}$ and $\mat{X}$. 
Given a matrix $\mat{X}$, the submatrix corresponding to the rows in set $R$ and columns in set $C$ is denoted $\mat{X}[R,C]$, and $\mat{X}[\cdot,C]$ or $\mat{X}[R,\cdot]$ when $R = \emptyset$ or $C = \emptyset$, respectively. 
We use parentheses and subscripts $(\mat{X})_{r,c}$ to denote the choice of single elements in the $r$ row and $c$ column of $\mat{X}$. 
We will occasionally drop the parentheses when such notation is superfluous from context. 
The normal distribution with mean $\vec{\mu}$ and covariance $\mat{\Sigma}$ is denoted $\mathcal{N}(\vec{\mu}, \mat{\Sigma})$. 

\section{Sheaf Neural Networks}

Sheaf neural networks were proposed by~\citet{hansen2020sheaf} as a generalization of graph convolutional networks~\cite{kipf2016semi} to cellular sheaf-structured data.
Rooted in topology and homological algebra, cellular sheaves are a natural object through which to view signals over graph structures that are subject to particular constraints on the data between neighboring nodes. 
This additional structure provided by cellular sheaves affords sheaf neural networks the ability to distinguish classes of graphs that GCNs cannot while tempering structural issues like oversmoothing~\cite{hansen2020sheaf, bodnar2022neural}.
We begin with a brief overview of cellular sheaves and their spectral properties before introducing sheaf neural networks. 

\subsection{Cellular Sheaves}

A \emph{cellular sheaf} is an algebro-topological data structure which associates a graph's nodes and edges to data in another space. 
Formally, a cellular
sheaf $\Fc$ on an undirected graph $G = (V,E)$ is specified by
\begin{itemize}
\item
  a vector space $\Fc(v)$ for each vertex $v \in V$
\item
  a vector space $\Fc(e)$ for each edge $e \in E$, and
\item
  a linear map $\mat{\Fc}_{v \face e} : \Fc(v) \rightarrow \Fc(e)$ for each incident vertex-edge
  pair $v \face e$ of $G$.
\end{itemize} Sheaves impose consistency constraints on the data assigned to incident vertices across edges through the restriction maps $\mat{\Fc}_{v \face e}$. 
Specifically, given an edge $e$ between vertices $u$ and $v$, we say that a choice of data $\vec{x}_v \in \Fc(v)$, $\vec{x}_u \in \Fc(u)$ is \emph{consistent} over $e$ if $\mat{\Fc}_{v \face e} \vec{x}_v = \mat{\Fc}_{u \face e} \vec{x}_u$. 
The product space of data associated with all vertices of $G$ is called the space of $0$-cochains and is denoted $C^0(G;\Fc)$.
For our purposes, we may view $C^0(G;\Fc)$ as a space of \emph{signals} on the vertices of $G$, where the value of a signal at a vertex $v$ lives in the vector space $\Fc(v)$. 
Similarly, we denote the space of signals associated with edges by $C^1(G;\Fc)$.
Each edge of $G$ imposes a constraint on $C^0(G;\Fc)$ by restricting the space associated with its two incident vertices. 
The subspace of $C^0(G;\Fc)$ satisfying all these constraints is the space of \emph{global sections} of $\Fc$, and is denoted $H^0(G;\Fc)$.
Data in $H^0(G'\Fc)$ are the assignments which satisfy the constraints introduced by $\Fc$ on $G$. 

The space of global sections $H^0(G;\Fc)$ is the kernel of a linear map $\mat{\delta}: C^0(G;\Fc) \to C^1(G;\Fc)$ called the \emph{coboundary}, and, given an arbitrary choice of orientation on the edges of the graph, may be computed by $$(\vec{\delta} \vec{x})_e = \mat{\Fc}_{v \face e} \vec{x}_v - \mat{\Fc}_{u \face e} \vec{x}_u$$ for each oriented edge $e = u \to v$. Therefore, if $\vec{x} \in \ker \mat{\delta}$, then $\mat{\Fc}_{v \face e} \vec{x}_v = \mat{\Fc}_{u \face e} \vec{x}_u$ for every edge $e = u \sim v$.

Sheaves have their own Laplacian operator~\cite{hansen2019toward} which simplifies to the graph Laplacian when the constraints imposed by the restriction maps are lifted such that $\mat{\Fc}_{v \face e} = \mat{I}$ for all $v \in V, e \in E$. 
The construction of the sheaf Laplacian mirrors the approach for the graph Laplacian as the matrix product of incidence matrices. 
Given a coboundary operator, the \emph{sheaf Laplacian} is given by $\mat{L}_\Fc = \mat{\delta}^T\mat{\delta}$, which is a positive semidefinite linear operator on $C^0(G;\Fc)$ with kernel $H^0(G;\Fc)$.

Rather than simply recording connections between nodes, cellular sheaves specify relationships between data associated with those nodes. Standard graph-theoretic constructions like Laplacians and diffusion operators implicitly work with the \emph{constant sheaf} on a graph: the sheaf $\constshf{\mathbb{R}}$ with all stalks $\mathbb{R}$ and all restriction maps the identity. This is a simple relationship between nodes which can be greatly generalized in the sheaf setting. For instance, a sheaf can easily represent a signed graph by changing the sign of one restriction map of the constant sheaf for each negatively signed edge. More general relationships between nodes can be expressed, especially as stalks increase in dimension, resulting in such operators as connection Laplacians \cite{singer_vector_2012} and matrix-weighted Laplacians \cite{tuna_synchronization_2016}.
These generalizations have just begun to be explored in the context of graph learning~\cite{hansen2020sheaf, bodnar2022neural}.

\subsection{Diffusion and Sheaf Convolutional Networks}

Graph convolutional networks~\cite{kipf2016semi} exploit local graph diffusion operators to define a neighborhood convolution operation over nodes, acting as a generalization of convolutional neural networks over an irregular domain defined by a graph.
These graph diffusion operations are typically implemented through a transformation of the graph's adjacency matrix $\mat{A}$ or, for GCNs in particular, the normalized Laplacian matrix $\mat{D}^{-\frac{1}{2}}\mat{A}\mat{D}^{-\frac{1}{2}}$ where $\mat{D}$ is the degree matrix of $G$.

Sheaf convolutional networks employ an analogous diffusion operation to define convolution-like operations of signals on $\Fc$. 
The sheaf Laplacian $\mat{L}_\Fc = \mat{\delta}^T\mat{\delta}$ encodes this diffusion operation in sheaf convolutional networks and, like the graph Laplacian, fulfills a number of desirable properties.
The zero eigenspace of $\mat{L}_{\Fc}$ corresponds to the sections of $\Fc$, meaning we may interpret this operator's spectral structure as providing information on the signals which represent  consistent assignments of data given the sheaf constraints.
For an appropriately chosen $\alpha$, the operator $\mat{Q}_\Fc^\alpha = \mat{I} - \alpha \mat{L}_\Fc$ will have $2$-norm 1 and has $H^0(G;\Fc)$ as the eigenspace corresponding to the eigenvalue $1$. 
There is also a normalized form of the sheaf Laplacian, $ \mat{\tilde{L}}_\Fc =\mat{D}^{-1/2}\mat{L}_\Fc \mat{D}^{-1/2}$, where $\mat{D}$ is the block diagonal of $\mat{L}_\Fc$. 
This normalization effectively reparameterizes the stalks of $\Fc$ so that the eigenvalues are bounded below by $0$ and above by $2$. Therefore, we can also construct a stable diffusion operator $\mat{\tilde Q}_\Fc = \mat{I} - \mat{\tilde{L}}_\Fc$. 
Diffusion operators depending on larger neighborhoods may be constructed from powers of these operators. 
For any $l$, $\tilde{\mat{Q}}_\Fc^l$ and $(\mat{Q}_\Fc^\alpha)^l$ are
$l$-step sheaf diffusion operators.

With this information, we can now define a sheaf convolutional layer. 
Assume our input data comes in the form of signals $\mat{X}$ over a sheaf $\Fc$.
As this is sheaf-structured data, these signals may be multi-dimensional such that each node $v \in V$ has a variable-dimensional stalk $\Fc(v)$.
For simplicity, we will assume each stalk is $k$-dimensional.
In addition, we assume each node signal has $d$ channels, forming a $d$-dimensional vector in $\mathbb{R}^d$.
When $k = 1$, $\mat{X}$ is a $(N_V \times d)$ matrix and assumes the form of the more familiar feature matrix in standard graph learning.
When $k > 1$, $\mat{X}$ has shape $N_V k \times d$ and contains $N_v$ blocks of $k$ rows which correspond to the $k$-dimensional stalk assignments of features over each node with the $d$ channels of each of these $k$-dimensional features comprising the columns. 

\begin{definition}
Let $\Df$ be some diffusion operator for the sheaf $\Fc$ such as $\Df = \mat{I} - \mat{\tilde{L}}_{\Fc}$. Given a choice of nonlinearity $\sigma$, a \emph{sheaf convolutional layer} for sheaf signals $\mat{X}_l$ in layer $l$ is defined as
\begin{equation*}
    \mat{X}_{l+1} = \sigma(\Df(\mat{I} \otimes \mat{B}_l)\mat{X}_l\mat{W}_l
\end{equation*} where $\mat{W}_l$ is a $(d_{l+1} \times d_l)$ weight matrix, $\mat{B}_l$ a $(k \times k)$ weight matrix, and $(\mat{I} \otimes \mat{B})$ the Kronecker product of an $(N_V \times N_V)$ identity matrix with $\mat{B}$. 
\end{definition}
The right multiplication of $\mat{W}_l$ with $\mat{X}_l$ enacts a linear transformation of the channels in an equivalent manner to traditional GCNs. The left multiplication of $(\mat{I} \otimes \mat{B}_l)$ allows for a linear transformation of each channel's feature vector by multiplying each row block of $\mat{X}_l$ by $\mat{B}_l$. When $\Fc$ is the constant sheaf, $k = 1$, and $\mat{B}_l$ the identity matrix, the sheaf convolutional layer corresponds to a traditional graph-convolutional layer. 
We may compose these sheaf convolutional layers to create a sheaf convolutional network with $L$ layers via
\begin{equation}\label{eq:sheaf_convolutional_network}
    f(\mat{W},\mat{B},G)(\mat{X}) =  \sigma(\Df (\mat{I} \otimes \mat{B}_L) \cdots \sigma(\Df (\mat{I} \otimes \mat{B}_2) \sigma(\Df(\mat{I} \otimes \mat{B}_1)\mat{X}\mat{W}_1)\mat{W}_2)\cdots \mat{W}_{L}).
\end{equation} Note this architecture is composed only of sheaf convolutional layers.
In graph classification tasks, it is common to add a ``readout" operation $\rho$ at the final layer which combines the latent node features to create an output classification or regression prediction for the task. 
A simple example of such a readout function is sum-aggregation $\rho(\mat{X}_L) = \sum_{i=1}^{N_V} (\mat{X}_L \vec{1})_v$.
In node classification tasks, the readout function may take the form of a sigmoid operation which maps the hidden node representations themselves to output class or regression values instead of aggregating over nodes~\cite{xu2018powerful}. 

\section{Neural Tangent Kernels}
We now turn our attention to graph neural tangent kernels and their extension to sheaves. 
After introducing the graph neural tangent kernel, we will derive a geometrically-focused parameterization of the neural tangent kernel for the sheaf convolutional network architecture given in the previous section. 

The neural tangent kernel~\cite{jacot2018neural} for fully-connected and convolutional networks describes the behavior and asymptotic performance of these networks under the assumption that they are trained by gradient descent with an infinitesimally small learning rate, are initialized randomly, and have layers of infinite width.
Under these assumptions, one can replace these infinitely-wide networks with a deterministic kernel machine whose kernel is given by 
\begin{equation*}
\mat{\Theta}(\vec{x},\vec{x}') = \mathbb{E}_{\mat{W} \sim \mathcal{N}(\vec{0}, \mat{I})}
                    \left[\left\langle \frac{\partial f(\mat{W},\vec{x})}{\partial \mat{W}},
                    \frac{\partial f(\mat{W},\vec{x}')}{\partial \mat{W}}
                    \right\rangle\right]
\end{equation*}
where $\vec{x}, \vec{x}'$ are two inputs and $f$ is a feedforward neural network with parameters $\mat{W}$.
\citet{du2019graph} showed that this notion of the neural tangent kernel extends to graph neural network architectures, and the authors provide a recursive formulation termed the graph neural tangent kernel. 
We will use this as our starting point for describing a geometrically-focused parameterization of the neural tangent kernel for sheaf neural networks and, by extension, graph convolutional networks.

\subsection{A Sheaf Neural Tangent Kernel}
Given a set of $n$ input sheaves $\{\Fc_i\}_{i=1}^n$, the supervised sheaf learning task looks to learn a sheaf convolutional network $f$ parameterized by weights $\mat{W}$ and $\mat{B}$ as structured in Equation~\ref{eq:sheaf_convolutional_network} with an additional readout function $\rho$. 
We would like to study the neural tangent kernel corresponding to this network, given by
\begin{equation*}
    \mat{\Theta}(\Fc, \Fc') = \mathbb{E}_{\mat{W} \sim \mathcal{N}(\vec{0}, \mat{I})}\left[\left\langle \frac{\partial f(\mat{W}, \Fc)}{\partial \mat{W}}, \frac{\partial f(\mat{W}, \Fc')}{\partial \mat{W}}\right\rangle\right]
\end{equation*} for two input cellular sheaves $\Fc, \Fc'$ based on graphs $G = (V,E)$ and $G' = (V', E')$ with the same number of nodes $|V| = |V'| = N_V$ but potentially differing number of edges.
In addition, we assume that $\Fc$ and $\Fc'$ have the same stalk dimensionality $k$, resulting in sheaf signals $\mat{X}, \mat{X}'$ of size $(N_v k \times d)$. 
Sheaves $\Fc$ and $\Fc'$ give rise to potentially distinct diffusion operators $\Df, \Df'$. 
Finally, we assume $\mat{B}$ and $\mat{B}'$ are fixed across layers of $f$ and simplify notation by integrating these constant matrices into the diffusion operators $\Df$ and $\Df'$, respectively. 
Taking the partial derivative with respect to the parameters in layer $l$, we see
\begin{equation*}
    \frac{\partial f(\mat{W}, \Fc)}{\partial \mat{W}_l} = \left(\Df\sigma(\vec{f}_l)\right)^\top \mat{P}_l
\end{equation*} where $\mat{P}_l = \Df^\top \mat{P}_{l+1} \mat{W}_{l+1}^\top \odot \dot{\sigma}(\vec{f}_l)$ captures the gradient propagating backwards from deeper network layers. 
Note the slight abuse of notation by viewing the bold-cased $\vec{f}_l$ as the vector-valued output of the $l$-layer sheaf convolutional network $f_l$.
Denoting the inner product $\mat{\theta}(f)$, we have
\begin{equation}\label{eq:full_inner_product}
\begin{split}
    \mat{\theta}(f) &= \left\langle \frac{\partial f(\mat{W}, \Fc)}{\partial \mat{W}}, \frac{\partial f(\mat{W}, \Fc')}{\partial \mat{W}}\right\rangle \\
    &= \left\langle \left(\Df\sigma(\vec{f}_l)\right)^\top \mat{P}_l, (\Df'\sigma(\vec{f}'_l))^\top \mat{P}_l' \right\rangle \\
    &= \left\langle \left(\Df\sigma(\vec{f}_l)\right)^\top, (\Df'\sigma(\vec{f}'_l))^\top \right\rangle \odot \left\langle \mat{P}_l, \mat{P}_l' \right\rangle \\
    &= \mat{\Sigma}_l \odot \langle \mat{P}_l, \mat{P}_l' \rangle.
\end{split}
\end{equation} Focusing on the left side of the Hadamard product in Equation~\ref{eq:full_inner_product}, we see a structure akin to the covariance of diffusion operations applied to the node representations at each layer. 
In the input layer, $\mat{\Sigma}_0 = \mat{X}\mat{X'}^\top$ is precisely the channel covariance of the input signals, and similarly, $\mat{\Sigma}_1 = \Df \mat{X}\mat{X}'^\top \Df'^\top$.
For layers $l > 1$, more care is required to write $\mat{\Sigma}_l$ in closed form due to the introduction of weights and nonlinearities in the message passing updates.
To arrive at a representation for $\mat{\Sigma}_l$, we will take expectations of layer-wise activations under the assumption that the network layers of $f$ are infinitely wide ($d_l \rightarrow \infty$) and weights are initialized according to a standard normal distribution $\mathcal{N}(\vec{0}, \mat{I})$. 
These assumptions lead to the following proposition.
\begin{proposition}\label{prop:sheaf_diffusion_covariance}
    Given an infinitely-wide sheaf convolutional network with parameters initialized according to a standard normal distribution, element-wise activation function $\sigma$, and sheaf signals $\mat{X}, \mat{X}'$ over sheaves with diffusion operators $\Df, \Df'$, the covariance of the diffusion operation gradients at layer $l$ is defined recursively as $$\mat{\Sigma}_l = \Df\mat{H}_{l-1}\Df'^\top$$ where
    $$\mat{H}_{l-1} = \mathop{\mathbb{E}}\limits_{_{\vec{f}_{l-1},\vec{f}_{l-1}' \sim \mathcal{N}(\vec{0}, \mat{\Sigma}_{l-1})}}\left[\sigma(\vec{f}_{l-1})\sigma(\vec{f}_{l-1}')^\top\right].$$
\end{proposition} The proof of this proposition follows from the recursive definition given in~\cite{du2019graph}, but a full derivation may be found in the appendix. 
Combining Proposition~\ref{prop:sheaf_diffusion_covariance} with the inner product between the deeper layers' backpropagated gradients according to Equation~\ref{eq:full_inner_product} results in a geometrically-oriented parameterization of the sheaf neural tangent kernel.
\begin{proposition}\label{prop:sheaf_neural_tangent_kernel}
    For an $L$-layer sheaf convolutional network structured as in Equation~\ref{eq:sheaf_convolutional_network} with fixed $\mat{B}$ at each layer, its corresponding neural tangent kernel between two sheaves $\Fc$ and $\Fc'$ may be written
    \begin{equation}\label{eq:sheaf_neural_tangent_kernel}
    \begin{split}
        \mat{\Theta}(\Fc, \Fc') &= \sum\limits_{l=1}^{L+1} \left(\mat{\Sigma}_l \odot (\Df \Df'^\top)^{\odot (L+1-l)}\right) \odot \left(\bigodot\limits_{i=l}^{L+1-l} \dot{\mat{H}}_i \right)  \\
        &= \sum\limits_{l=1}^{L+1} \mat{\Delta}_l \odot \mat{\Pi}_l
    \end{split}
    \end{equation}
\end{proposition} where $\dot{\mat{H}}_l = \mathbb{E}_{\vec{f}_l,\vec{f}_l' \sim \mathcal{N}(\vec{0}, \mat{\Sigma_l})}[\dot{\sigma}(\vec{f}_l) \dot{\sigma}(\vec{f}_l')^\top]$, $\mat{\Delta}_l = \mat{\Sigma}_l \odot (\Df \Df'^\top)^{\odot (L+1-l)}$, and $\mat{\Pi}_l = \bigodot_{i=l}^{L+1-l} \dot{\mat{H}}_i$. 
The proof of this proposition may be found in the appendix and aligns with previous work on graph convolutional networks when the sheaf stalk dimensionality $k=1$~\cite{sabanayagam2021new}. Note that in the node classification setting $\Df = \Df'$.

Equation~\ref{eq:sheaf_neural_tangent_kernel} clearly delineates the structure of $\mat{\Theta}$ as the sum of element-wise products of diffusion-related effects $\mat{\Delta}_l$ with parameter path activation effects $\mat{\Pi}_l$.
To further clarify this relationship, consider the case when the activation function $\sigma$ is the identity. 
Under this scenario, each $\dot{\mat{H}}_i$ is $\mat{1}$, and the tangent kernel becomes
\begin{equation}\label{eq:linear_sheaf_neural_tangent_kernel}
    \begin{split}
        \bar{\mat{\Theta}}(\Fc, \Fc') &= \sum\limits_{l=1}^{L+1} \Df^{l}\mat{X}\mat{X}'^\top (\Df'^{\top})^{l} \odot (\Df \Df'^\top)^{\odot (L+1-l)} \\
        &= \sum\limits_{l=1}^{L+1}\bar{\mat{\Delta}}_l.
\end{split}
\end{equation} 
The neural tangent kernel for such a linear sheaf convolutional network lacks the dependence on the activation paths and takes the form of a weighted sum over products of the $l$-step diffusion of signals on sheaves $\Fc$ and $\Fc'$. 

\section{Discussion}
The derivation of the sheaf neural tangent kernel given in Equation~\ref{eq:sheaf_neural_tangent_kernel} produces a number of intuitive insights which assist in the interpretation of both the function and limitations of sheaf convolutional networks and, by extension, graph convolutional networks. 

\subsection{Relationship to Graph Convolutional Networks}
When the input space is structured such that $\Fc, \Fc'$ are both constant sheaves with $1$-dimensional stalks ($k = 1$), Equation~\ref{eq:sheaf_neural_tangent_kernel} is equivalent to the graph neural tangent kernel between graphs $G$ and  $G'$ with diffusion matrices given by, for example, adjacency matrices $\mat{A}, \mat{A}'$ or their transformations as respective graph Laplacians 
$\Df = (\mat{I} - \mat{D}^{-\frac{1}{2}}\mat{A}\mat{D}^{-\frac{1}{2}})$, $\Df' = (\mat{I} - \mat{D}'^{-\frac{1}{2}}\mat{A}'\mat{D}'^{-\frac{1}{2}})$~\cite{sabanayagam2021new}. 
In other words, graph convolutional networks operate on sheaves with trivial structure. 
This also implies $\bar{\mat{\Theta}}$ for trivial sheaf structures is closely related to the neural tangent kernel of the simplified graph convolutional network architecture described in~\citet{wu2019simplifying}. 
These \emph{simple graph convolutional networks} are structured as $$f^{\text{smp}}(\mat{W},G, L)(\mat{X}) = \rho(\Df^L \mat{X} \mat{W}).$$
Therefore, the corresponding neural tangent kernel is given by $$\bar{\mat{\Theta}}_L^{\text{smp}} = \Df^L\mat{X}\mat{X}'^\top(\Df'^\top)^L \odot (\Df\Df'^\top).$$
The neural tangent kernel for linear graph convolutional networks is composed of weighted sums of $\bar{\mat{\Theta}}_l^{\text{smp}}$ for each layer $l$.
This close approximation is reflected by the empirical results of ~\citet{wu2019simplifying} which show that simple graph convolutional networks can achieve performance in line with those of more complicated graph neural network architectures on particular tasks. 
These results provide some confidence that even our linearized sheaf neural tangent kernel may still provide insight into the behavior of the more complex graph convolutional network architectures used in practice. 

\subsection{Oversmoothing}
As observed by~\citet{bodnar2022neural}, the trivial structure imposed by GCNs is intimately related to the tendency of these architectures to bias towards learning over-smoothed, homophilic representations of the input graph signals. 
Our neural tangent kernel derivation provides further confirmation of the presence of these biases in GCNs. 

Equation~\ref{eq:linear_sheaf_neural_tangent_kernel} shows that, for a network of depth $L$, the kernel value $\bar{\mat{\Theta}}(G,G')$ between graph signals $\mat{X}, \mat{X}'$ on graphs $G$ and $G'$ will contain terms consisting of signals whose feature values have been diffused across $\{1, 2, \dots, L\}$ steps along the graph. 
Although the deeper diffusion terms are exponentially down-weighted according to the element-wise power of $\Df \Df'^\top$, their oversmoothing effects on deep graph convolutional networks are worrisome due to the fact that $\lim_{L \rightarrow \infty} \Df^L$ will approach an orthogonal projection onto the harmonic space $H^0(G,\constshf{\Fc})$.
This subspace consists of constant functions on $G$ which lack discriminative power over the nodes of $G$. 
As a result, $\mat{\Theta}(G,G')$ for a sufficiently deep graph convolutional network will approximate the inner product of the projection of signals $\mat{X}$ and $\mat{X}'$ onto the kernel of $\Df$ and $\Df'$. 
For exceptionally deep networks, we can view $\bar{\mat{\Theta}}(G,G')$ as returning the similarity between the limiting distributions of random walks on $G$ and $G'$. 
Spectral bounds for such random walk processes are well-studied~\cite{chung1997spectral, lovasz1993random}, and the incorporation of such methods in analyzing the discriminatory power of graph neural tangent kernels may prove insightful for future work.

We also view the homophilic bias imparted by the trivial diffusion structure of GCNs through this spectral lens. 
To see this, assume the setup of a 2-class node classification task such that the class assignments $\vec{y} \in \{-1,1\}$  partition of the nodes of $G$.
We can bound the mixing time $M(G)$ of $G$ by  
\begin{equation}\label{eq:mixing_bound}
    M(G) \sim O\left(\frac{\log\min_v(\vec{\pi}(G))_v^{-1}}{h(G)^2}\right)
\end{equation} where $\vec{\pi}(G)$ the limiting distribution of $G$ and $h(G)$ is the Cheeger constant:
\begin{equation*}
    h(G) = \min\limits_{S} \left\{\frac{|\partial S|}{\min\{|S|,|\overline{S}|\}}\right\}
\end{equation*} where $S \subset G$ and $\partial S$ is the edge boundary of $G$ composed of the edges which connect nodes in $S$ and $\overline{S}$. 
The bound in Equation~\ref{eq:mixing_bound} is maximized for homophilic graphs which are composed of a small number of densely-connected clusters with weak between-cluster connectivity.
Such graphs have small $h(G)$ and will mix more slowly over a fixed number of diffusion steps, leading to increased separation in the tangent kernel between nodes of different classes for $L$ fixed (assuming the class assignments $\vec{y}$ respect the homophilic structure).

The bound on $M(G)$ is minimized when $G$ is a complete bipartite graph, causing the diffusion process executed by a graph convolutional network to approach harmonicity at an even faster rate and decreasing the separability of the neural tangent kernel. 
Worse, when $G$ is a connected bipartite graph and the distribution of class labels is opposite across two equal node partitions of $G$, the diffusion process reaches a steady state immediately, sending the node representations to the kernel of $\Df$ and trivializing $\bar{\mat{\Theta}}$. 
By contrast, consider the same same assumptions on $G$, but with $\Df$ composed of restriction maps with opposing signs on each incident edge $\Fc_{v \face e} = -\Fc_{u \face e}$.
Signals diffusing over this sheaf oscillate across nodes in each bipartition instead of being immediately sent to the kernel of $\Df$. 

These observations, which align with the results on the linear separability of sheaf diffusion discussed in~\citet{bodnar2022neural}, reveal that the constraints imposed by sheaf structures are crucial to learning over particular graph structures when applying convolutional architectures. 
Through the addition of proper sheaf constraints, one can ameliorate the tendency of graph convolutional networks to mix too quickly over particular graph structures and consequently learning simplistic node representations.
Unfortunately, the necessary constraints are typically unknown \textit{a priori}.
Although it may be possible to learn these constraints from raw graph signals themselves~\cite{bodnar2022neural}, the question of which sheaf constraints are optimal for a given input graph structure remains an important open question.

\subsection{Relationship to Diffusion Kernels}
As noted in the previous section, the linear sheaf neural tangent kernel $\bar{\mat{\Theta}}$ is determined by the weighted sum of diffusion-like operations.
The use of diffusion as a graph kernel is a well-studied topic in the traditional graph kernel literature~\cite{smola2003kernels}.
Using this graph kernel language, the structure of $\bar{\mat{\Theta}}$ within each layer $l$ may be described as the composition of two $l$-step random walk kernels $\mat{K} = (\alpha\mat{I} - \tilde{\mat{L}})$~\cite{kondor2002diffusion} and an inner product kernel. 
This relationship offers an interesting avenue for future work in determining the extent to which neural network architectures may be augmented such that their neural tangent kernels approximate compositions of more exotic graph kernels. 

\subsection{Influence of Parameter Connectivity}
Our discussion thus far has focused on the diffusion terms $\mat{\Delta}_l$ of of the tangent kernel as given in Equation~\ref{eq:sheaf_neural_tangent_kernel}. 
However, the effects of parameter paths as encoded by $\mat{\Pi}_l$ on  $\mat{\Theta}$ cannot be ignored, especially when $\sigma(x) = \max\{0,x\}$. 

With ReLU activation functions, each $\mat{\Pi}_l$ acts as a mask on the diffusion kernel values in $\mat{\Delta}_l$, zeroing diffusion kernel values between nodes $u$ and $v$ for if their weighted activation becomes negative at any proceeding layer.
In this way, $\mat{\Theta}$ is the element-wise product of two descriptions of connectivity between $u$ and $v$: one coming from the similarity of an $l$-step diffusion of their signals along the graph, and the other the weighted connectedness of their features through $\mat{W}$.

Under the neural tangent kernel assumptions, $\mat{W}$ is normally distributed and does not change during training. 
Abusing this model slightly, we can hypothesize that finite-width networks will adjust $\mat{W}$ to weight distinctly features resulting from diffusion operations at different layers.
In other words, ReLU activations allow the network to control which steps of the random walk it attends to given an input signal on the graph, thereby learning the important degrees of locality for a task. 
Although in practice it appears that neural tangent kernels can approximate the behavior of finite-width networks~\cite{lee2020finite}, more work is required to show how this approximation behaves as layer width decreases and how this movement into the feature regime affects our understanding of graph and sheaf neural tangent kernels as being driven by diffusion processes on the underlying network. 

\bibliography{icml2022.bib}
\bibliographystyle{iclr2022_workshop}
\appendix
\section{Proof of proposition~\ref{prop:sheaf_diffusion_covariance}}
For $l > 1$ we wish to compute $$\mat{\Sigma}_l = \langle \left(\Df\sigma(\vec{f}_l)\right)^\top, (\Df'\sigma(\vec{f}'_l))^\top \rangle.$$
Let us focus on the value $\mat{\Sigma}[v,v']_{\kappa, \kappa'}$ which corresponds to the inner product between feature dimensions $\kappa, \kappa'$ of node features of $v,v' \in V$:
\begin{equation*}
    \left\langle (\Df\sigma(\vec{f}_{l-1}))[v,\cdot], (\Df'\sigma(\vec{f}'_{l-1}))[v',\cdot] \right\rangle_{\kappa, \kappa'} = \sum\limits_{j=1}^{d_{l-1}}\left( (\Df\sigma(\vec{f}_{l-1}))[v,j]_\kappa \right) \left((\Df'\sigma(\vec{f}'_{l-1}))[v',j]_{\kappa'} \right).
\end{equation*} Assuming $d_{l-1} \rightarrow \infty$, we may reinterpret the above as an expectation over the feature dimension:
\begin{equation*}
    \lim\limits_{d_{l-1} \rightarrow \infty} \sum\limits_{j=1}^{d_{l-1}}\left( (\Df\sigma(\vec{f}_{l-1}))[v,j]_\kappa \right) \left((\Df'\sigma(\vec{f}'_{l-1}))[v',j]_{\kappa'} \right) = \mathbb{E}\left[(\Df\sigma(\mat{f}_{l-1}))[v,j]_{\kappa} (\Df'\sigma(\vec{f}_{l-1}))[v',j]_{\kappa'} \right].
\end{equation*} We seek to separate the message passing effects from the activation-related terms.
Unpacking the interior matrix multiplication and rearranging terms we have
\begin{align*}
    \mathbb{E}\left[(\Df\sigma(\mat{f}_{l-1}))[v,j]_{\kappa} (\Df'\sigma(\vec{f}_{l-1}))[v',j]_{\kappa'} \right] &= \mathbb{E}\left[\left(\sum\limits_{t \in v} (\Df[v,t])_\kappa(\sigma(\vec{f}_{l-1})[t,j])_{\kappa'}\right) \left(\sum_{s \in V}(\Df'[v',s])_{\kappa}(\sigma(\vec{f}'_{l-1})[s,j])_{\kappa'} \right) \right] \\
    &= \mathbb{E}\left[\sum\limits_{t\in V}\sum\limits_{s \in V} (\Df[v,t])_{\kappa}(\Df'[v',s])_\kappa' (\sigma(\vec{f}_{l-1})[t,j])_k (\sigma(\vec{f}'_{l-1})[s,j])_{\kappa'} \right] \\
    &= \sum\limits_{t\in V}\sum\limits_{s \in V} (\Df[v,t])_{\kappa} \mathbb{E}\left[(\sigma(\vec{f}_{l-1})[t,j])_k (\sigma(\vec{f}'_{l-1})[s,j])_{\kappa'} \right](\Df^\top[s,v'])_{\kappa'} \\
    &= \Df[v,\cdot]_\kappa \mat{H}_{l-1}[\kappa, \kappa'] (\Df[v',\cdot]_{\kappa'})^\top
\end{align*} We can further roll this into the submatrix of all feature covariances between $v$ and $v'$ as
\begin{align*}
    \left\langle (\Df\sigma(\vec{f}_{l-1}))[v,\cdot], (\Df'\sigma(\vec{f}'_{l-1}))[v',\cdot] \right\rangle &= \Df[v',\cdot]\mat{H}_{l-1}(\Df[v',\cdot])^\top \\
    &= \mat{\Sigma}_l[v,v']
\end{align*} where $\mat{H}_{l-1} = \mathop{\mathbb{E}}\limits_{\vec{f}_{l-1}, \vec{f}'_{l-1} \sim \mathcal{N}(\vec{0}, \mat{\Sigma}_{l-1})}[\sigma(\vec{f}_{l-1}) \sigma(\vec{f}_{l-1})^\top]$. \qed

\section{Proof of proposition~\ref{prop:sheaf_neural_tangent_kernel}}
We would like to compute the inner product between feature dimensions $\kappa$ and $\kappa'$ of $\mat{P}_l[v,\cdot]$ and $\mat{P}'_l[v',\cdot]$ for nodes $v$ and $v'$. 
We have
\begin{equation*}
    \begin{split}
        \left\langle \mat{P}_l[v,\cdot], \mat{P}'_l[v',\cdot] \right\rangle_{\kappa,\kappa'} &= \mat{P}_l[v,\cdot]_\kappa (\mat{P}'_l[v',\cdot]_{\kappa'})^\top \\
        &= \left((\Df^\top \mat{P}_{l+1}\mat{W}_{l+1}^\top \odot \dot{\sigma}(\mat{f}_l)) (\Df'^\top \mat{P}'_{l+1}\mat{W}_{l+1}^\top \odot \dot{\sigma}(\mat{f}'_l))^\top\right)[v,v']_{\kappa,\kappa'} \\
        &= \sum\limits_{j=1}^{d_l} \left(\Df^\top \mat{P}_{l+1}\mat{W}_{l+1}^\top \odot \dot{\sigma}(\mat{f}_l)\right)[v,j]_{\kappa} \left(\Df'^\top \mat{P}'_{l+1}\mat{W}_{l+1}^\top \odot \dot{\sigma}(\mat{f}'_l)\right)[v',j]_{\kappa'} \\
        &= \sum\limits_{j=1}^{d_l}\left(\sum\limits_{t \in V} \sum\limits_{i=1}^{d_{l+1}} (\Df[t,v])_{\kappa}(\mat{P}_{l+1}[t,i])_{\kappa}(\mat{W}_{l+1})_{j,i}(\sigma(\mat{f})_{l}[v,j])_{\kappa}\right) \\
        &\cdot \left(\sum\limits_{s \in V'} \sum\limits_{h=1}^{d_{l+1}} (\Df'[s,v'])_{\kappa'}(\mat{P}'_{l+1}[s,h])_{\kappa'}(\mat{W}_{l+1})_{j,h}(\sigma(\mat{f}')_{l}[v',j])_{\kappa'}\right) \\
        &= \sum\limits_{i=1}^{d_{l+1}} \sum\limits_{h=1}^{d_{l+1}} ((\Df^\top \mat{P}_{l+1})[v,i]_k (\Df'^\top \mat{P}'_{l+1})[v',h]_{\kappa'} \sum\limits_{j=1}^{d_l} (\mat{W}_{l+1}^2)_{j,i} \dot{\sigma}(\mat{f}_l)[v,j]_\kappa\dot{\sigma}(\mat{f}'_{l})[v',j]_{\kappa'}. 
    \end{split}
\end{equation*}

Again letting $d_l \rightarrow \infty$, we know that $\mathbb{E}[\mat{W}^2_{l+1}] = 0$ for off-diagonal terms under our i.i.d. normal Gaussian assumptions on the initialization of $\mat{W}$. 
The diagonal of $\mat{W}$ is the variance which is $1$ under these assumptions. 
This allows us to simplify the above in the limit as
\begin{equation*}
\begin{split}
\left\langle \mat{P}_l[v,\cdot], \mat{P}'_l[v',\cdot] \right\rangle_{\kappa,\kappa'} &= \left\langle (\Df^\top \mat{P}_{l+1})[v,\cdot]_{\kappa}, (\Df'^\top \mat{P}'_{l+1})[v',\cdot]_{\kappa} \right\rangle \mathbb{E}[\dot{\sigma}(\mat{f}_{l})[v,j]_{\kappa} \dot{\sigma}(\mat{f}'_{l})[v',j]_{\kappa'}] \\
&= \left\langle (\Df^\top \mat{P}_{l+1})[v,\cdot]_{\kappa}, (\Df'^\top \mat{P}'_{l+1})[v',\cdot]_{\kappa} \right\rangle \dot{\mat{H}_l}[v,v']_{\kappa, \kappa'} \\
&= \left\langle (\Df [v,\cdot]_{\kappa}, (\Df' [v',\cdot]_{\kappa} \right\rangle \odot \left\langle \mat{P}_{l+1}[v,\cdot]_{\kappa}, \mat{P}'_{l+1}[v',\cdot]_{\kappa} \right\rangle \dot{\mat{H}_l}[v,v']_{\kappa, \kappa'}
\end{split}
\end{equation*}


Finally, the expectation of the inner product $\mat{\theta}(f)[v,v']_{\kappa,\kappa'}$ along nodes $v$ and $v'$ in features $\kappa$, $\kappa'$ may be written
\begin{equation*}
    \begin{split}
        \mathop{\mathbb{E}}_{\mat{W} \sim \mathcal{N}(\vec{0}, \mat{I})}\mat{\theta}(f)[v,v']_{\kappa,\kappa'} &= \mathop{\mathbb{E}}_{\mat{W} \sim \mathcal{N}(\vec{0}, \mat{I})}\left[\left\langle \left(\frac{\partial f(\mat{W}, \Fc)}{\partial \mat{W}}\right)[v,\cdot], \left(\frac{\partial f(\mat{W}, \Fc')}{\partial \mat{W}}\right)[v',\cdot]\right\rangle_{\kappa, \kappa'} \right] \\
        &= \left\langle (\Df\sigma(\vec{f}_{l-1}))[v,\cdot]_{\kappa}, (\Df'\sigma(\vec{f}'_{l-1}))[v',\cdot]_{\kappa'} \right\rangle
        \odot \left\langle \mat{P}_l[v,\cdot]_{\kappa}, \mat{P}'_l[v',\cdot]_{\kappa'} \right\rangle \\
        &= \mat\Sigma_l[v,v']_{\kappa, \kappa'}(\Df \Df^\top)[v,v']_{\kappa, \kappa'} \left\langle \mat{P}_{l+1}, \mat{P}'_{l+1} \right\rangle[v,v']_{\kappa, \kappa'} \dot{\mat{H}}_l[v,v']
    \end{split}
\end{equation*} which we can fill in recursively given our definition of $\left\langle \mat{P}_l[v,\cdot], \mat{P}'_l[v',\cdot] \right\rangle_{\kappa,\kappa'}$ above, resulting in 
\begin{equation*}
    \begin{split}
    \mathop{\mathbb{E}}_{\mat{W} \sim \mathcal{N}(\vec{0}, \mat{I})}\mat{\theta}(f)[v,v']_{\kappa,\kappa'} &= \mathop{\mathbb{E}}_{\mat{W} \sim \mathcal{N}(\vec{0}, \mat{I})}\left[\left\langle \left(\frac{\partial f(\mat{W}, \Fc)}{\partial \mat{W}}\right)[v,\cdot], \left(\frac{\partial f(\mat{W}, \Fc')}{\partial \mat{W}}\right)[v',\cdot]\right\rangle_{\kappa, 
    \kappa'} \right] \\
    &= \mat{\Sigma}_{l}[v,v']_{\kappa, \kappa'} ((\Df \Df^\top)[v,v']_{\kappa, \kappa'})^{L+1-l}\left(\prod\limits_{i=l}^{L+1-l}\dot{\mat{H}}_l[v,u]_{\kappa, \kappa'} \right).
    \end{split}
\end{equation*} Expanded to all features of all nodes and summing across layers in accordance to \citet{du2019graph}, we have
\begin{equation*}
    \begin{split}
        \mat{\Theta}(\Fc, \Fc') &= \sum\limits_{l=1}^{L+1} \mat{\Sigma}_l \odot (\Df \Df'^\top)^{\odot (L+1-l)} \odot \left(\bigodot\limits_{i=l}^{L+1-l} \dot{\mat{H}}_i \right)
\end{split}
\end{equation*} as desired. \qed 

\end{document}